\documentclass[10pt,twocolumn,letterpaper]{article}

\usepackage[pagenumbers]{wacv}

\usepackage{amsmath,amssymb,booktabs}
\usepackage{colortbl}
\usepackage{caption}
\definecolor{ctrlrow}{gray}{0.92}

\definecolor{wacvblue}{rgb}{0.21,0.49,0.74}
\usepackage[pagebackref,breaklinks,colorlinks,allcolors=wacvblue]{hyperref}

\title{Future Rendering $\neq$ Future Surface: A Benchmark and Dataset\\
for Dynamic Surface Reconstruction Beyond the Observed Window}

\author{%
Yukun Shi \qquad Minglun Gong\\
University of Guelph
}

\begin{document}
\makeatletter
\twocolumn[{%
  \@maketitle
  \centering
  \includegraphics[width=\linewidth]{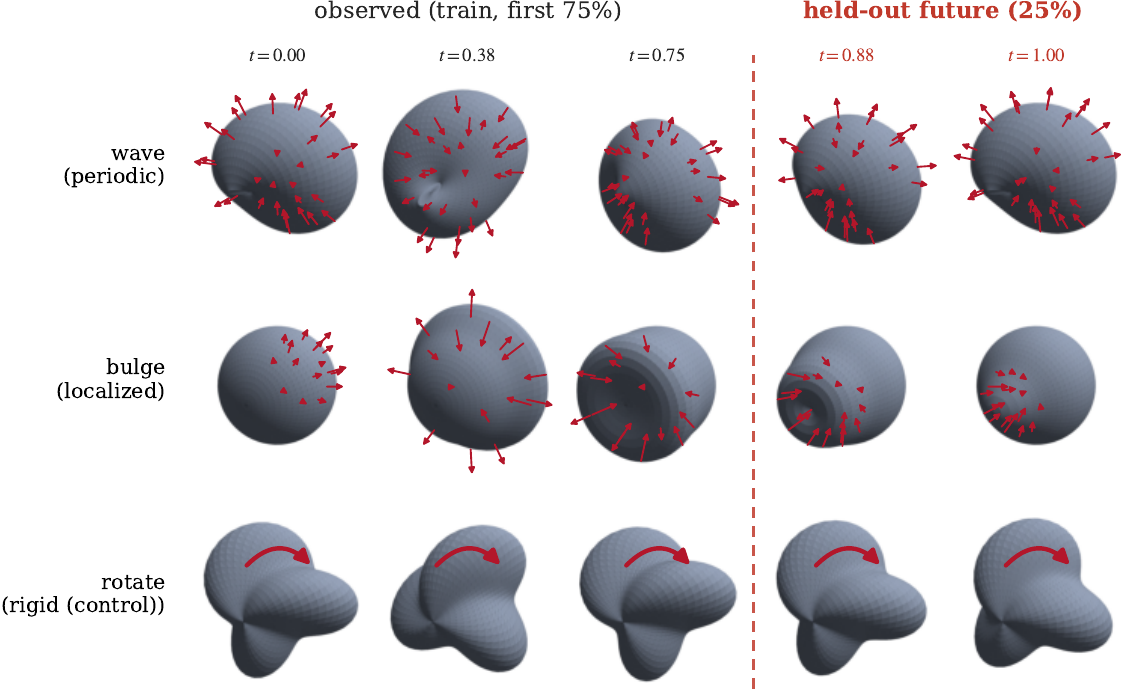}
  \captionof{figure}{\textbf{\textsc{FutureSurf}: does the reconstructed \emph{surface} survive beyond
  the observed window?} Three of our eight controlled motions, shown as ground-truth mesh sequences
  (red arrows: ground-truth surface velocity), isolate periodic motion, localized deformation, and a
  rigid-motion control. A method trains on the first $75\%$ of timestamps
  (\emph{observed}) and is scored on its extracted surface over the held-out \emph{future} ($25\%$).
  Because these futures are analytically defined, the benchmark can test whether a learned dynamic
  reconstruction carries the surface into held-out time. We release the dataset, splits, scorer, and
  ground-truth-side recoverability oracle.}
  \label{fig:teaser}
  \vspace{8pt}
}]
\makeatother

\begin{figure*}[t]\centering
\includegraphics[width=\linewidth]{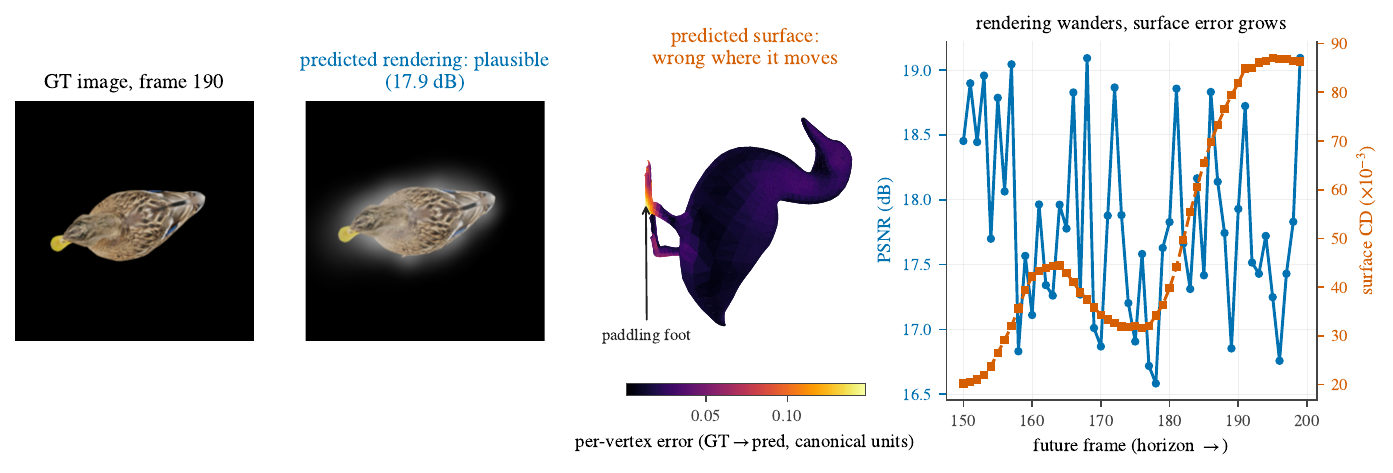}
\caption{\textbf{Future rendering $\neq$ future surface} (duck, DG-Mesh, far-horizon future frame 190).
\emph{Left pair:} the predicted rendering remains a recognizable duck (PSNR $=17.9$\,dB).
\emph{Middle:} the reconstructed \emph{surface} at the same frame is substantially wrong where the
surface moves (per-vertex error peaks on the paddling foot). \emph{Right:} over the whole future
window, per-frame rendering quality stays in a narrow band while per-frame surface error grows
several-fold. At frame~190 the Chamfer distance (CD) is $81.8\times10^{-3}$, about $4\times$ the
observed-window error; lower CD is better. The two errors are weakly rank-correlated
(duck Spearman $\rho(\mathrm{LPIPS},\mathrm{CD})=+0.09$), so rendering metrics do not track future geometry.}
\label{fig:hero}
\end{figure*}

\begin{abstract}
Dynamic-scene reconstruction is almost always evaluated inside the \emph{observed} time window, yet deployment settings such as AR overlays, robot interaction, and anticipatory planning need the \emph{future} surface: the geometry at times beyond those captured. No standard benchmark measures this. We introduce \textsc{FutureSurf}, a \emph{controlled diagnostic} benchmark and dataset for future-time surface reconstruction that trades scene diversity for exact future ground truth and falsification controls. A method trains on the observed first $75\%$ of a sequence; we score its extracted per-frame surface on the held-out future by Chamfer distance, reporting absolute future CD as the primary score and the future/observed \emph{gap} as a diagnostic. The dataset contains eight analytically defined controlled motions, including three falsification controls, with \emph{exact} per-frame ground-truth meshes. We also provide a ground-truth-side recoverability oracle. The release includes split files, CPU scoring code, a benchmark card, and Croissant metadata. On the controlled motions, the DG-Mesh backbone leaves a $2.7\text{--}4.1\times$ gap even for futures predictable \emph{in principle} (four of five recoverable from observed motion by a fixed rule), while the falsification controls behave as designed (the surface-invariant motion shows no gap). Beyond the contributed dataset, the gap persists across six animated DG-Mesh asset scenes and a second backbone, Deformable-3DGS ($2.0\text{--}6.6\times$; both share a deformation-MLP temporal model). The benchmark also shows that future rendering quality and future-surface accuracy are statistically \emph{decoupled} (mean per-frame rank correlation $|\rho(\mathrm{LPIPS},\mathrm{CD})|\!=\!0.13$), so the novel-view-synthesis metrics the field reports do not track future geometry. The future error is structured, concentrating where the surface moves. The dataset, evaluation toolkit, and scoring code are publicly available on Hugging Face (\url{https://huggingface.co/datasets/rickyshi/futuresurf}) and GitHub (\url{https://github.com/Ricky-S/futuresurf}).
\end{abstract}

\section{Introduction}
Reconstructing dynamic 3D scenes from video has become a central problem, with deformable Gaussian and neural-field methods achieving high-fidelity novel-view synthesis and, increasingly, temporally consistent surfaces. Yet the standard evaluation protocol measures quality \emph{within} the observed time window: it asks how well a model interpolates the times it was trained on. Many downstream uses are fundamentally about the \emph{future}: an AR system must place content on a surface a moment before it is observed; a robot must anticipate where a deforming object will be; a planner must reason about geometry it has not yet seen. For these, the relevant question is \emph{future-surface} accuracy: given frames up to time $T$, how good is the reconstructed surface \emph{mesh} at times $t>T$ (Fig.~\ref{fig:hero})?

This question is, to our knowledge, unmeasured. Future-time work on dynamic Gaussians forecasts \emph{appearance} and reports rendering metrics (PSNR/SSIM/LPIPS); dynamic-surface work reports Chamfer or F-score but only \emph{within} the observed window. The gap is concrete: \emph{no standardized protocol evaluates the reconstructed future surface mesh at held-out times}, so the field lacks a controlled way to study when and why future-surface reconstruction fails.

We close this gap with an \emph{evaluation contribution}. \textsc{FutureSurf} (Fig.~\ref{fig:teaser}) packages a held-out-future split and per-frame Chamfer metric in a shared canonical frame, a controlled dataset of analytically defined motions with per-frame ground-truth meshes and falsification controls, and a benchmark card stating what claims the benchmark can and cannot support. Everything needed to evaluate a new method (data, splits, and scoring scripts) is provided.

\paragraph{Contributions.}
\begin{enumerate}\itemsep0pt
\item A \textbf{dataset and benchmark for future-time surface reconstruction}: eight analytically defined controlled motions (three of them falsification controls) with \emph{exact} per-frame ground-truth meshes, a held-out-future split, a method-agnostic Chamfer protocol with released scoring code, a benchmark card, and Croissant metadata. It is the first standardized testbed that scores the reconstructed future \emph{surface mesh} at held-out times, where prior future-geometry work scores points or renderings rather than extracted meshes (\S\S\ref{sec:dataset},\,\ref{sec:protocol}).
\item The \textbf{future-surface gap} the benchmark exposes: even on the constructed motions, whose futures are predictable \emph{in principle} (four of five recoverable from observed motion by a fixed rule), DG-Mesh leaves a large gap ($2.7\text{--}4.1\times$); the gap also appears on the asset scenes across two deformation-MLP backbones ($2.0\text{--}6.6\times$), while the falsification controls behave as designed (\S\ref{sec:gap}).
\item A \textbf{rendering--surface decoupling} result: the NVS metrics the field reports (PSNR, LPIPS) do not track future-surface accuracy, so future-rendering quality is not sufficient evidence of future-surface quality (\S\ref{sec:diag}).
\end{enumerate}

\textsc{FutureSurf} is a \emph{controlled diagnostic} benchmark. Where large dataset releases~\cite{kubric} maximize scene diversity, it provides \emph{exact future ground truth, falsification controls, and a recoverability oracle}: the instrumentation existing benchmarks lack to isolate surface accuracy at held-out future times. It thus supports claims about \emph{relative} future-surface accuracy, motion-type dependence, and the rendering--surface decoupling, and explicitly does \emph{not} claim to cover all dynamic scenes or to certify real-world deployment. Synthetic, analytically defined motion is necessary: it is the only way to obtain exact per-frame ground-truth meshes \emph{together with} the controls and oracle. The gap and predictability structure are verified on \emph{both} backbones; the rendering--surface decoupling and the gauge decomposition are reported on DG-Mesh, whose surface extraction supports them.

\section{Related Work}\label{sec:related}
We organize prior work by \emph{what it evaluates}, since that is where our gap lies.

\paragraph{Future-time dynamic scenes (appearance).} A growing line of work forecasts dynamic Gaussians or radiance fields to \emph{future} times: latent-ODE extrapolators~\cite{odegs}, Gaussian-prediction forecasters~\cite{gaussianprediction}, learned dynamical-system models over 4D Gaussians~\cite{evogs}, Gaussian-process temporal extrapolation with uncertainty~\cite{gp4dgs}, and motion-aware Gaussian grouping for space-time forecasting~\cite{mogaf}. All of these evaluate future \emph{renderings} (typically PSNR/SSIM/LPIPS), never the future surface mesh; our decoupling result (\S\ref{sec:diag}) shows the two are not interchangeable.

\paragraph{Dynamic surface reconstruction (within-window).} Building on Gaussian-splatting and deformation-field representations~\cite{threedgs,twodgs,fourdgs,dnerf,hypernerf}, a line of methods reconstructs dynamic \emph{surfaces} and reports geometry metrics (Chamfer, F-score, or accuracy/completion)~\cite{dynasurfgs,spacetime2dgs,gaustar,foursurf,dysurface}, but only on the \emph{observed} window: they measure interpolation, not extrapolation. We evaluate two dynamic-Gaussian backbones \emph{beyond} the observed window: DG-Mesh~\cite{dgmesh} (a surface method) and Deformable-3DGS~\cite{deformable3dgs} (a deformation-MLP backbone). Extracting \emph{consistent, per-frame surface meshes} from monocular video is itself a nascent capability, and DG-Mesh is among the few methods that do so, which is why we instantiate the surface-extraction backbone with it. Because the protocol scores only \emph{exported per-frame meshes}, independent of how a method produces them, it can benchmark any method in this emerging class (not just DG-Mesh) and will apply to new methods as they appear.

\paragraph{Future geometry and gauge.} A physics-prior line couples reconstruction with a physical simulator to advance or synthesize future dynamics~\cite{springgaus,physgaussian,gic,fluidnexus}, but reports particle/point or rendering error under an assumed physical model rather than a standardized surface-mesh protocol. The closest effort to our setting is ReconPhys~\cite{reconphys}, which predicts Gaussian-center \emph{point sets} from a single video via a feed-forward physics model, evaluated as points rather than extracted surface meshes, with no standardized surface-mesh protocol or diagnostic. The use of explicit dynamics priors in this line is consistent with our empirical finding (\S\ref{sec:gap}) that the tested deformation-MLP backbones do not reliably carry surface geometry into held-out future time without additional temporal inductive bias. Separately, feed-forward 3D predictions are often evaluated after removing scale/alignment (gauge) ambiguity~\cite{dust3r}; our diagnostic explicitly separates this removable gauge component from the non-rigid core. Table~\ref{tab:compare} situates the gap this benchmark fills.

\begin{table}[t]\centering\scriptsize
\setlength{\tabcolsep}{2.5pt}\renewcommand{\arraystretch}{1.1}
\begin{tabular}{@{}p{0.34\linewidth}ccccc@{}}
\toprule
 & future & surface & per-frame & protocol & diagn.\,$+$ \\
 & time & mesh & GT mesh & $+$\,suite & controls \\
\midrule
within-window surface \cite{dgmesh,dynasurfgs,spacetime2dgs,gaustar,foursurf,dysurface} & -- & \checkmark & partial & -- & -- \\
future-time appearance \cite{odegs,gaussianprediction,evogs,gp4dgs,mogaf} & \checkmark & -- & -- & -- & -- \\
ReconPhys~\cite{reconphys} & \checkmark & points & -- & -- & -- \\
\textsc{FutureSurf} (ours) & \checkmark & \checkmark & \checkmark & \checkmark & \checkmark \\
\bottomrule
\end{tabular}
\caption{What existing evaluation practice measures vs.\ this benchmark. No prior work scores the
reconstructed \emph{surface mesh} at held-out \emph{future} times under a standardized, released
protocol with diagnostics and falsification controls.}
\label{tab:compare}
\end{table}

\paragraph{Benchmark practice.} Our packaging follows established dataset-documentation practice: the benchmark card is in the spirit of datasheets~\cite{datasheets} and model cards~\cite{modelcards}, and the dataset ships with machine-readable Croissant metadata~\cite{croissant}.

\section{The \textsc{FutureSurf} Dataset}\label{sec:dataset}
\textsc{FutureSurf} provides per-frame ground-truth \emph{surface} meshes at future times: the data the task needs and real captures cannot supply. It comprises \textbf{eight constructed controlled motions}, where the future is analytically known and predictable in principle. All eight share one capture convention: a per-vertex-textured surface under a monocular orbit camera, $200$ frames, the first $75\%$ (frames $0$--$149$) observed and the last $25\%$ ($150$--$199$) held out for future scoring. Every frame includes an \emph{exact} ground-truth mesh (Tab.~\ref{tab:stats}).

\paragraph{Why controlled synthetic data.} Measuring future-\emph{surface} accuracy requires an \emph{exact} ground-truth mesh at future times. A real capture yields future \emph{images}, but not the true surface; obtaining that surface would require another reconstruction step with its own error. Animated or analytic assets with per-frame meshes are therefore required by the task itself, which helps explain why this quantity has not become a standard metric. We use this requirement deliberately: each constructed motion isolates a temporal factor or superposition, the controls expose the protocol itself, and futures predictable in principle separate ``the future is unknowable'' from ``the tested backbone does not extrapolate.'' This follows established precedent that controlled synthetic data is a first-class evaluation instrument~\cite{kubric}, and that protocol-level reality checks need not introduce a new method to be a contribution~\cite{dycheck}.

\begin{table}[t]\centering\footnotesize
\setlength{\tabcolsep}{6pt}\renewcommand{\arraystretch}{1.05}
\begin{tabular}{@{}ll@{}}
\toprule
motion & temporal factor isolated \\
\midrule
wave & periodicity \\
compound & superposition (periodicity $+$ trend) \\
stretch & monotonic trend \\
bulge & spatial locality (traveling bump) \\
accel & acceleration \\
\midrule
\rowcolor{ctrlrow} twist$^\dagger$ & surface-invariant motion \\
\rowcolor{ctrlrow} rotate$^\dagger$ & pure gauge (rigid spin) \\
\rowcolor{ctrlrow} stop$^\dagger$ & static future (frozen at $t{=}0.5$) \\
\bottomrule
\end{tabular}
\caption{Our contributed dataset: eight constructed motions, each with $1{,}986$ ground-truth vertices,
$200$ frames ($150/50$ holdout split), per-frame ground-truth meshes, and Croissant metadata
(Fig.~\ref{fig:dataset}). Five isolate surface-changing temporal factors (top); three are falsification controls
($^\dagger$, bottom).}
\label{tab:stats}
\end{table}

\paragraph{The eight controlled motions.} Each constructed motion deforms a common base UV sphere (radius $0.9$) by an analytic, time-parameterized map $F_m(x,t)$, so the future surface is known in closed form at every $t$ and the challenge is carried by the motion's \emph{time} profile, not its static shape (several motions are near-indistinguishable in any single frame and differ only in how they evolve). Five \emph{surface-changing} motions each isolate a temporal factor or superposition: \emph{wave}, a pure periodic radial ripple that returns to its start (periodicity); \emph{compound}, the same ripple superimposed on a monotonic elongation, so it never returns (superposition); \emph{stretch}, a smooth anisotropic elongation on a decelerating speed profile (monotonic trend); \emph{bulge}, a traveling Gaussian protrusion sweeping in polar angle (spatial locality); and \emph{accel}, the \emph{identical} elongation as stretch but on a constant-acceleration profile still speeding up at the split (acceleration). Stretch and accel are a deliberate pair: they share one static shape but opposite speed profiles, so a constant-velocity guess overshoots one and undershoots the other. No single per-vertex extrapolation rule covers the suite, which spans multiple rule families (\S\ref{sec:baselines}). The generator equations for all eight maps are in the supplementary; as concrete examples, with $n$ the unit radial direction and $\theta,\phi$ the azimuth and polar angle, the \emph{wave} and \emph{bulge} maps are
\[
\begin{aligned}
\text{wave}&: x + 0.22\sin(2\pi t + 1.2\phi + \theta)\,n,\\
\text{bulge}&: x + 0.34\exp\!\big(\!-(\phi-\pi t)^2/(2\!\cdot\!0.28^2)\big)\,n.
\end{aligned}
\]

\paragraph{Three falsification controls.} Three further motions expose the \emph{protocol} rather than any method, so a broken metric or alignment cannot pass silently. \emph{twist} (surface-invariant) is an azimuthal shear that displaces vertices substantially (mean $0.24$, max $0.47$) while leaving the \emph{surface} almost unchanged (surface change $0.014$): a correct protocol must report a unit gap, separating surface change from parameterization motion. \emph{rotate} (pure gauge) is a rigid spin of a fixed asymmetric shape, whose designed future motion is a removable global similarity transform; it calibrates the gauge decomposition. \emph{stop} (static future) is the wave frozen at $t{=}0.5$, so the held-out future is \emph{static and already observed}: the correct prediction is ``do nothing,'' and the metric exposes whether a backbone drifts anyway.

\paragraph{Additional third-party evaluation scenes.} Beyond the contributed dataset, and only for shape diversity and a cross-backbone check, we also \emph{evaluate on} six animated asset scenes from DG-Mesh~\cite{dgmesh} (duck, horse, girlwalk, torus$\to$sphere, bird, beagle).

\begin{figure}[t]\centering
\includegraphics[width=0.88\linewidth]{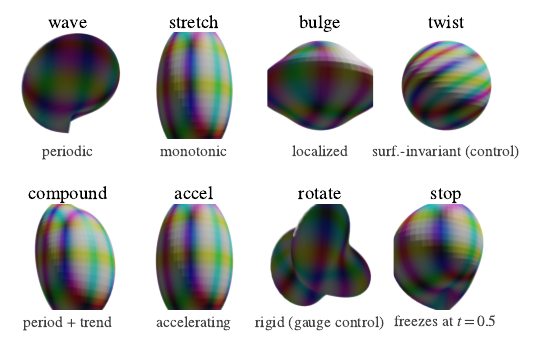}
\caption{The controlled-motion dataset (representative frames): eight analytically defined motions under the identical
capture convention (per-vertex-textured surface, shared camera trajectory, per-frame ground-truth
meshes). Five surface-changing motions isolate different temporal factors; three falsification controls test the protocol.}
\label{fig:dataset}
\end{figure}

\paragraph{Benchmark card and metadata.} A datasheet-style benchmark card is included in the release. \emph{Intended use:} a controlled diagnostic benchmark for measuring and diagnosing future-time surface reconstruction under identical splits. \emph{Supported claims:} relative future-surface accuracy across methods; motion-type dependence of extrapolation difficulty; the rendering--surface decoupling. \emph{Unsupported claims:} real-world deployment performance (the data is synthetic) and appearance quality as a primary benchmark target. \emph{Assumptions:} per-frame ground-truth meshes, monocular orbit capture, and a shared canonical frame for prediction and ground truth (no per-frame alignment). \emph{Responsible use:} the constructed data is procedurally generated (no personal or human-subject data). The constructed motions are released under CC\,BY\,4.0 with Croissant metadata; the benchmark is versioned and new motions are added as separate scene sets.

\section{Task, Protocol, and Metric}\label{sec:protocol}
Table~\ref{tab:protocol} summarizes the protocol; we detail it below.
\begin{table}[t]\centering\footnotesize
\setlength{\tabcolsep}{3pt}\renewcommand{\arraystretch}{1.15}
\begin{tabular}{@{}p{0.24\linewidth}p{0.72\linewidth}@{}}
\toprule
\textbf{Task} & given frames $t\in[0,T]$, reconstruct the surface mesh at held-out $t>T$ \\
\textbf{Split} & first $75\%$ of timestamps train; remaining $25\%$ evaluate \\
\textbf{Ground truth} & per-frame ground-truth meshes \\
\textbf{Metric} & \emph{primary:} absolute per-frame future CD (canonical frame, GT scale) \\
\textbf{Diagnostics} & gap $=$ future/observed CD (mean/median); Sim(3) gauge-removed CD (oracle); per-vertex error map \\
\textbf{Artifacts} & dataset, split files, scoring scripts, benchmark card \\
\bottomrule
\end{tabular}
\caption{\textsc{FutureSurf} protocol at a glance. A method is evaluated by training on the
observed window and scoring its \emph{extracted meshes} at held-out future times; all scoring is
script-driven and method-agnostic.}
\label{tab:protocol}
\end{table}

\textbf{Task and split.} Given frames $t\in[0,T]$, reconstruct the surface mesh at held-out $t>T$; the first $75\%$ of timestamps train, the remaining $25\%$ evaluate. \textbf{Metric.} Per-frame bidirectional Chamfer between the \emph{vertex sets} of the extracted mesh and the ground-truth mesh, expressed in a shared \emph{canonical frame} (one fixed coordinate frame at ground-truth scale, with no per-frame alignment). Chamfer matches each vertex to its nearest neighbor on the other mesh, so scoring needs no vertex correspondence between the predicted and ground-truth meshes and no shared topology: a method is judged only on where its surface lies. The \emph{primary score} is absolute future CD, directly comparable across methods at a fixed scene; the same qualitative conclusions hold under a surface-sampled Chamfer check that does not depend on mesh vertex density.

\textbf{Diagnostics.} The \emph{gap} (future CD / observed-window CD) compares a method's future-window error (extrapolation) to its observed-window error (interpolation). It is the paper's central \emph{diagnostic}, reported throughout, but not the primary score: as a ratio, it inherits run-to-run variance from its denominator. The Sim(3)-aligned CD (``gauge-removed'': best per-frame rotation, translation, and uniform scale) is an \emph{oracle diagnostic}, used to separate removable global drift from non-rigid surface error.

\paragraph{Formal definitions.} With $p,g$ ranging over the predicted and ground-truth vertex sets $P_t,G_t$ at frame $t$ (canonical GT-scale frame), the per-frame score is the symmetric Chamfer distance
\begin{equation}
d(P_t,G_t)=\tfrac12\!\operatorname*{mean}_{p}\min_{g}\|p\!-\!g\|
+\tfrac12\!\operatorname*{mean}_{g}\min_{p}\|g\!-\!p\| .
\label{eq:cd}
\end{equation}

\begin{figure}[t]\centering
\includegraphics[width=\linewidth]{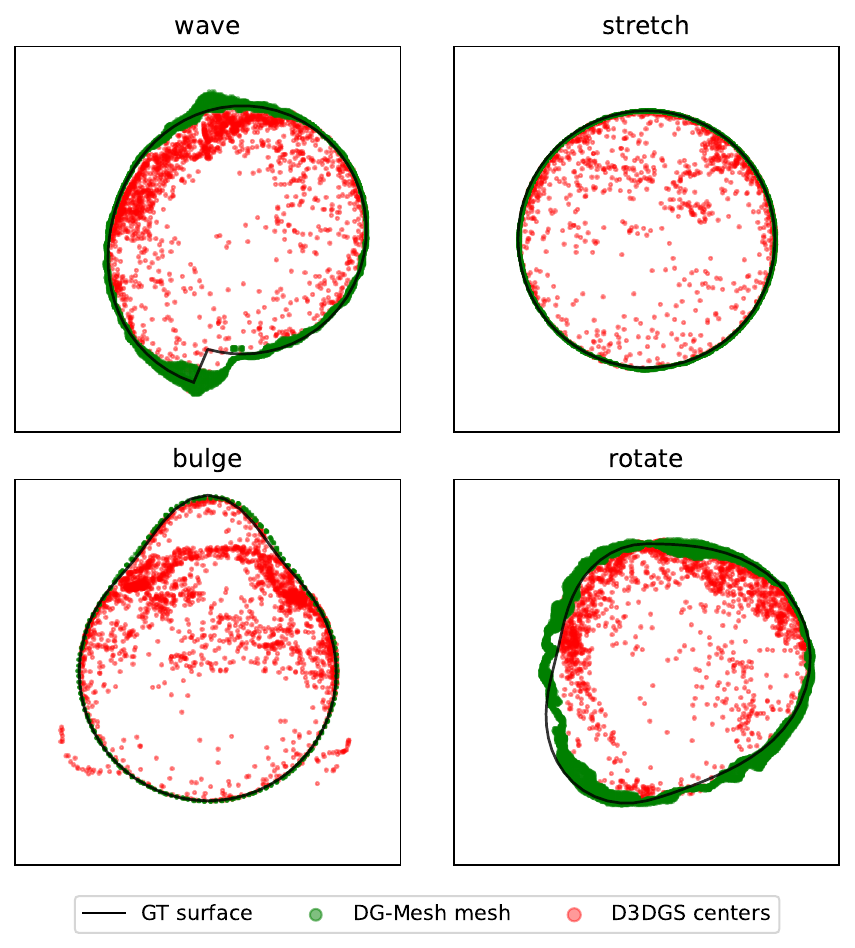}
\caption{\textbf{On the analytic surfaces, the second backbone is a point cloud, not a surface mesh.}
Cross-sections of four constructed motions: DG-Mesh's mesh (green) lies on the ground-truth surface
(black); Deformable-3DGS's Gaussian centers (red) form a loose cloud $2\text{--}7\times$
farther from the surface than DG-Mesh's across all eight scenes (median over observed frames).}
\label{fig:offsurface}
\end{figure}

\paragraph{Design principles.} \emph{D1 (method-agnostic):} a method is evaluated only through its extracted per-frame meshes; nothing in the protocol assumes a representation. \emph{D2 (exact ground truth):} per-frame GT meshes come from animation or analytic definition, never from a sensor or a fitted proxy. \emph{D3 (falsifiable):} the suite contains motions designed to expose the \emph{protocol} itself (surface-invariant, rigid, frozen-future), so a broken metric or alignment cannot pass silently. \emph{D4 (diagnosable):} every result comes with a breakdown, not just a single number: how much of the error is a removable pose/scale offset versus genuine shape error (gauge vs.\ non-rigid), and a per-vertex map of where the error concentrates. A failure can therefore be \emph{explained}, not merely measured. \emph{D5 (observed-window precondition):} ``bad on the future'' is only meaningful once a method is \emph{good on the observed window}, so before scoring its future we require its observed-window CD to sit at GT scale: on the order of the surface backbone's observed-window CD on the same scenes, not several-fold above it. The reported observed-window CDs (Tab.~\ref{tab:master}, Tab.~\ref{tab:controlled}) confirm the evaluated backbones meet this; a method whose observed reconstruction sits off this scale (Deformable-3DGS's center cloud on the constructed surfaces, Fig.~\ref{fig:offsurface}) is out of scope for the future claim there.

\paragraph{Implementation and compute.} Each backbone is trained with its own released schedule (DG-Mesh extracts a mesh per frame; Deformable-3DGS uses its D-NeRF schedule); the benchmark changes nothing in training and is purely an evaluation protocol. Scoring is CPU-only: given per-frame meshes, the controlled-motion gap and recoverability numbers are produced by the released scripts in minutes.

\paragraph{Why the controlled study uses DG-Mesh.} Deformable-3DGS does not export a surface mesh: it outputs per-frame Gaussian \emph{point clouds}, while the benchmark scores a mesh. On the textured asset scenes its centers lie close enough to the surface to serve as a proxy, so we report it in the cross-backbone check (\S\ref{sec:gap}). On the smooth analytic constructed surfaces, however, the centers form a loose cloud around the surface, sitting $2\text{--}7\times$ further from it than DG-Mesh's mesh on every scene (Fig.~\ref{fig:offsurface}). Scoring this cloud as a surface is unreliable, so the controlled study, which needs an accurate surface for the exact ground-truth controls, uses DG-Mesh.

\section{The Future-Surface Gap}\label{sec:gap}
\begin{table}[t]\centering\footnotesize\setlength{\tabcolsep}{3pt}
\begin{tabular}{@{}lrrrrrr@{}}
\toprule
Scene & Obs CD & Fut CD & Raw gap & \textbf{Non-rigid} & PSNR & $\rho_{\text{L,CD}}$ \\
\midrule
duck & 0.0205 & 0.0497 & 2.4$\times$ & \textbf{1.9$\times$} & 17.8 & +0.09 \\
horse & 0.0152 & 0.0580 & 3.8$\times$ & \textbf{2.8$\times$} & 18.5 & -0.11 \\
girlwalk & 0.0148 & 0.0565 & 3.8$\times$ & \textbf{3.7$\times$} & 21.0 & -0.10 \\
torus$\to$sphere & 0.0165 & 0.1099 & 6.6$\times$ & \textbf{5.1$\times$} & 13.5 & -0.21 \\
bird & 0.0178 & 0.0917 & 5.2$\times$ & \textbf{4.8$\times$} & 17.8 & -0.14 \\
beagle & 0.0207 & 0.0414 & 2.0$\times$ & \textbf{1.4$\times$} & 17.0 & -0.16 \\
\bottomrule
\end{tabular}

\caption{Per-scene future-surface gap on the six DG-Mesh asset scenes. Obs/Fut CD are the observed/future-window absolute per-frame bidirectional Chamfer (canonical GT units); Raw gap $=$ Fut/Obs CD ($2.0\text{--}6.6\times$, from unrounded means); the gauge-removed (\textbf{non-rigid}) gap stays large (genuine surface error, not removable drift); PSNR is future-window rendering quality; $\rho_{\text{L,CD}}\!=\!\rho(\text{LPIPS},\text{CD})$ is weak (Spearman), showing rendering quality does not predict surface quality (rendering--surface decoupling).}
\label{tab:master}
\end{table}
\paragraph{Accurate observed reconstruction, degraded future surface.} The evaluated methods meet our observed-window precondition (D5: they reconstruct the observed surface at GT scale), yet their extracted future surfaces degrade sharply. We first establish the phenomenon on the six DG-Mesh asset scenes, which span diverse shapes: DG-Mesh future-surface error is $2.0\text{--}6.6\times$ its observed-window error (Tab.~\ref{tab:master}). The same pattern appears on the second backbone, Deformable-3DGS, whose asset-scene Gaussian centers stay close enough to the surface to serve as a proxy: its observed-window CD is small ($0.012\text{--}0.026$ in canonical GT units, unlike the analytic surfaces of Fig.~\ref{fig:offsurface}), ruling out gross misalignment or an already-failed observed-window reconstruction as the explanation for the future increase. Its gaps are then comparable ($2.1\text{--}6.6\times$; per-scene values in the supplement). These asset scenes provide shape and backbone breadth; the deeper diagnostics below (falsification controls, recoverability oracle, factor isolation) need exact ground truth, which only our controlled dataset provides.

\begin{figure}[t]\centering
\includegraphics[width=0.85\linewidth]{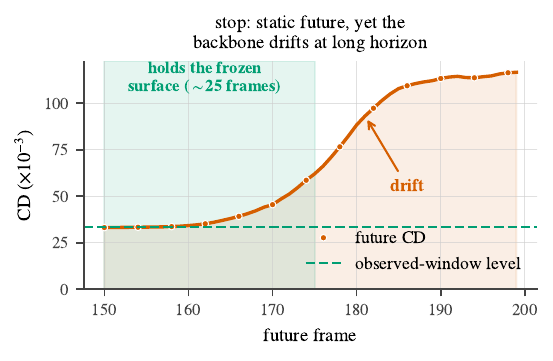}
\caption{The \emph{stop} control over the prediction horizon. With the motion frozen from
$t{=}0.5$, per-frame future CD stays within ${\sim}2\times$ the observed-window level for
${\sim}25$ frames, then drifts: the deformation field extrapolates non-constantly even though the
observed motion has already stopped.}
\label{fig:controls}
\end{figure}

\paragraph{The controlled motions isolate the extrapolation failure.} On the eight constructed motions (\S\ref{sec:dataset}), whose futures are simple and predictable \emph{in principle}, DG-Mesh confirms the gap: the method reconstructs the observed surface accurately yet extrapolates these futures poorly (Tab.~\ref{tab:controlled}).
\begin{table}[t]\centering\footnotesize\setlength{\tabcolsep}{4.5pt}
\begin{tabular}{llrrr}
\toprule
motion & character & obs CD & fut CD & gap \\
\midrule
wave & periodic & $0.0350$ & $0.0945$ & $2.7\times$ \\
compound & period $+$ trend & $0.0278$ & $0.0854$ & $3.1\times$ \\
stretch & monotonic & $0.0228$ & $0.0695$ & $3.0\times$ \\
bulge & localized & $0.0286$ & $0.0918$ & $3.2\times$ \\
accel & accelerating & $0.0210$ & $0.0861$ & $4.1\times$ \\
\midrule
\rowcolor{ctrlrow} twist$^\dagger$ & surf.-invariant & $0.0173$ & $0.0173$ & $1.0\times$ \\
\rowcolor{ctrlrow} rotate$^\dagger$ & rigid & $0.0348$ & $0.1189$ & $3.4\times$ \\
\rowcolor{ctrlrow} stop$^\dagger$ & freezes at $t{=}0.5$ & $0.0335$ & $0.0703$ & $2.1\times$ \\
\bottomrule
\end{tabular}
\caption{Controlled-motion results on our contributed dataset (DG-Mesh backbone): absolute per-frame bidirectional Chamfer (GT scale; lower is better) on the observed and future windows, and the gap
(future/observed CD; $1$ means the future surface is as accurate as the observed window, higher is worse).
The five surface-changing motions are listed first; the three $^\dagger$\,controls follow.}
\label{tab:controlled}
\end{table}
Despite small observed-window CDs, the five surface-changing motions all show large future gaps: $2.7\text{--}4.1\times$ by mean and $3.2\text{--}4.6\times$ by median, across periodic, period$+$trend, monotonic, localized, and accelerating motions. These futures are analytically defined, hence knowable by construction; the next section measures which ones a ground-truth-side oracle recovers. The gaps thus point to a limitation of the tested backbone's temporal extrapolation rather than intrinsic unknowability of these constructed futures. $^\dagger$\,The three \emph{controls} (\S\ref{sec:dataset}) behave as designed. \emph{twist}'s unit gap confirms the protocol measures surface change, not parameterization motion (it displaces vertices by $0.24$ on average while the surface itself changes by only $0.014$). \emph{rotate}'s future error is correctly attributed by the gauge decomposition to a removable global transform. \emph{stop} freezes the motion at $t{=}0.5$, so the correct prediction is ``do nothing'': the backbone holds the frozen surface for ${\sim}25$ frames in the headline run (mean gap $2.1\times$), then drifts from the static surface at longer horizons. Fig.~\ref{fig:controls} shows the \emph{stop} control over the horizon.

\section{Recoverability of the Constructed Futures}\label{sec:baselines}
\paragraph{Ground-truth-side oracle.} To turn ``predictable in principle'' into a measurement, we run the extrapolator suite (four naive rules, two learned families) directly on the \emph{ground-truth} vertex trajectories of the five surface-changing motions, scoring predicted future vertices against the future ground-truth surface (Tab.~\ref{tab:gtside}). These trajectories are well defined only because each controlled motion deforms a fixed-topology template, so a ground-truth vertex keeps its identity across frames. Our Chamfer metric does not need this correspondence (\S\ref{sec:protocol}): it matches nearest points between meshes, not tracked vertices. This is deliberately an \emph{oracle} evaluation, unavailable to image-based methods: it is an optimistic recoverability reference for the tested temporal rule families, not a baseline method.
\begin{table}[t]\centering\footnotesize\setlength{\tabcolsep}{4.5pt}
\begin{tabular}{llrr}
\toprule
motion & best rule & fut.\ CD & vs.\ freeze \\
\midrule
wave & harmonic $K{=}2$ & $0.0006$ & $47\times$ \\
compound & harmonic $K{=}2$ & $0.0014$ & $15\times$ \\
stretch & cubic & $0.0008$ & $13\times$ \\
accel & quadratic & $0.00001$ & ${\sim}1200\times$ \\
\midrule
bulge & ${\approx}$freeze (velocity) & $0.0173$ & $1.0\times$ \\
\bottomrule
\end{tabular}
\caption{Ground-truth-side (\emph{oracle}) recoverability of the constructed futures: future-oracle best
rule fit on the observed ground-truth vertex trajectories and scored against the future
ground-truth surface. CD is bounding-box-normalized linear Chamfer over vertex sets, on a different scale from Tab.~\ref{tab:controlled}'s absolute mesh CD; ratio vs.\ the do-nothing freeze reference. Four of five
futures are recovered nearly exactly; the traveling bump (bulge) defeats every tested per-vertex rule.}
\label{tab:gtside}
\end{table}
Four of the five constructed futures are recovered nearly exactly by simple rule families matched to their analytic construction, so the backbone's $2.7\text{--}4.1\times$ gaps occur on futures \emph{demonstrably recoverable from observed motion alone}. The exception is instructive: \emph{bulge} is a transport phenomenon (a bump traveling across the surface), and no per-vertex temporal rule meaningfully improves on the do-nothing freeze (best: velocity, $1.3\%$ below it). Its future is determined by construction, but recovering it would need a spatial-transport rule outside the per-vertex families our oracle tests. The constructed motions thus span rule \emph{families} as well as motion types. Wherever oracle recoverability is established, the future follows from the observed motion itself; the remaining gap reflects limitations of the tested backbones' representations and temporal extrapolation rather than intrinsic unknowability.

Simple observed-window extrapolation does not reliably solve the asset scenes either: held-out selection can miss the scene-dependent future-optimal rule, and the learned per-vertex model does not consistently beat the naive rules (full results in the supplement). \textsc{FutureSurf} is thus neither impossible nor solved by a single trivial temporal extrapolator.

\section{Diagnostics: Rendering--Surface Decoupling and Error Structure}\label{sec:diag}
\paragraph{Rendering $\neq$ surface.} Per-frame future RGB error and surface error are statistically decoupled across the six scenes (Tab.~\ref{tab:master}): their rank correlation is weak (mean $|\rho(\mathrm{LPIPS},\mathrm{CD})|\!=\!0.13$) and a linear fit explains under 6\% of the surface-error variance. The novel-view-synthesis metrics the field reports therefore do not track future-surface accuracy: future-rendering quality is not sufficient evidence of future-surface quality.

\begin{figure}[t]\centering
\includegraphics[width=0.8\linewidth]{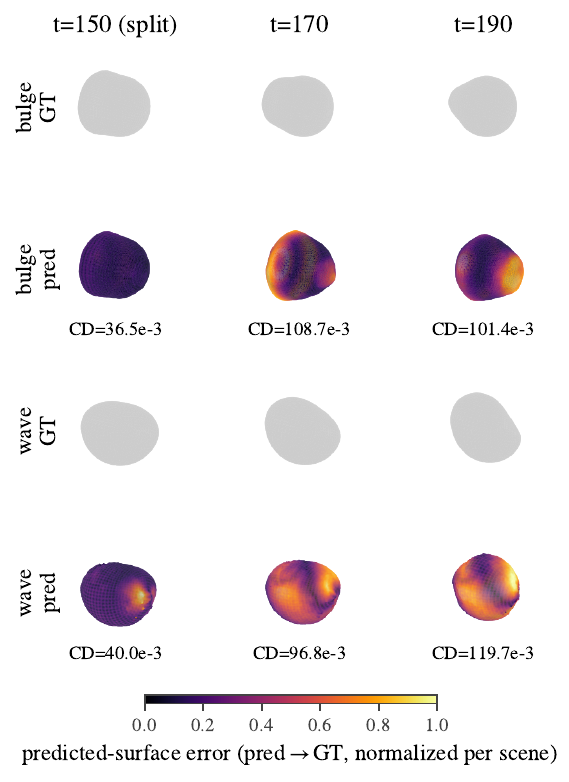}
\caption{Future-surface degradation over the prediction horizon (two constructed
motions, \emph{bulge} and \emph{wave}; DG-Mesh). For each: the GT mesh (gray) and
the reconstruction colored by per-vertex prediction-to-GT distance at the split frame $t{=}150$ and two later frames; per-frame CD annotated. Error grows sharply
over the future window and concentrates where the surface moves most (the bulge's traveling
protrusion, the wave's crests).}
\label{fig:strip}
\end{figure}

\paragraph{Where the error lives.} Future error is not spread evenly: it concentrates where the surface moves, growing with local motion on every scene and peaking at fast-moving features such as the bulge's traveling protrusion and the wave's crests (Fig.~\ref{fig:strip}). The same concentration pattern holds for Deformable-3DGS on the asset scenes (supplement). The remaining gap is genuine non-rigid shape error, not a removable global pose or scale offset: an oracle Sim(3) alignment leaves most of the deforming-scene gap intact (a non-rigid gap of $1.4\text{--}5.1\times$, Tab.~\ref{tab:master}), while on the rigid control it removes a far larger share, as expected when the motion is genuinely rigid (per-scene correlations, motion ratios, and the gauge decomposition in the supplement).

\section{Discussion and Limitations}
\paragraph{Limitations.} Ground-truth future geometry requires synthetic assets with per-frame meshes; we evaluate on the constructed suite plus the DG-Mesh asset collection, with a dataset-agnostic protocol released for extension. Capture follows the monocular-orbit convention. The protocol is method-agnostic, but breadth is bounded by which methods expose per-frame meshes: on the constructed suite Deformable-3DGS's Gaussian centers form a cloud off the surface (Fig.~\ref{fig:offsurface}), so scoring it as a surface is unreliable and the controlled study uses the surface-extraction backbone. Both backbones moreover share a time-conditioned deformation-MLP, so our evidence is limited to that family; a temporally distinct representation (physics-based, explicit 4D-grid, or neural-SDF flow) is a natural next backbone to separate the family's bias from the broader difficulty. The default split is 75/25, with horizon-dependence in per-frame curves; the gap persists at 60/40 and 85/15, and the scorer supports arbitrary splits. Finally, this paper diagnoses the gap rather than proposing a new reconstruction method: correcting the global gauge did not close it, consistent with error that is mostly non-rigid and scene-structured.

\paragraph{Implications for method design.} Future-surface metrics should accompany deployment claims: PSNR/LPIPS alone do not reliably track future geometry (\S\ref{sec:diag}). The results point to temporal extrapolation, not observed-window fit, as a bottleneck: simple per-vertex rules and learned per-vertex models do not reliably predict the future, and model selection from observed frames can choose the wrong rule. Future methods likely need stronger temporal inductive bias, such as dynamics priors, periodicity, physical constraints, or regularization against drift on static futures.

\paragraph{Artifact availability.} The released benchmark comprises the controlled-motion dataset (CC\,BY\,4.0, Croissant metadata), split files, evaluation code, and the benchmark card, available on \href{https://huggingface.co/datasets/rickyshi/futuresurf}{Hugging Face} and \href{https://github.com/Ricky-S/futuresurf}{GitHub}. The GT-side recoverability numbers reproduce \emph{without retraining or a GPU} from the released ground truth via \path{gt_oracle.py}; the DG-Mesh gap numbers reproduce by regenerating per-frame predictions with the public DG-Mesh code on the released dataset and scoring with \path{score.py}. Asset-scene results use the same protocol on the public DG-Mesh assets. The benchmark is versioned and maintained, with new motions added as backward-compatible scene sets so existing splits and reported numbers stay valid.

\section{Conclusion}
Future rendering $\neq$ future surface. The gap is large and structured by motion type on both tested deformation-MLP backbones; on the surface backbone it is decoupled from rendering quality, and on the constructed suite it points to how that family extrapolates over time, not to futures that cannot be predicted. We release \textsc{FutureSurf}, a controlled dataset and diagnostic benchmark, to make future-surface prediction measurable across representations.

\clearpage
{ \small
    \bibliographystyle{ieeenat_fullname}
    \bibliography{references}
}

\end{document}